\documentclass[10pt,twocolumn,letterpaper]{article}

\usepackage{cvpr}
\usepackage{times}
\usepackage{epsfig}
\usepackage{graphicx}
\usepackage{amsmath}
\usepackage{amssymb}
\usepackage{enumitem}
\usepackage{adjustbox}
\usepackage{tabularx}
\usepackage{stmaryrd}
\usepackage{mathtools}
\usepackage{float}
\usepackage{subcaption}
\usepackage[table]{xcolor}
\usepackage{tabu,multirow}
\usepackage{hhline}
\usepackage{url}
\usepackage{flushend}
\usepackage[mathcal]{euscript}

\renewcommand{\paragraph}[1]{\medskip\noindent\textbf{#1}\enskip}

\newcommand{\coco}{{\tt MS-COCO}}

\newcommand{\si}[2]{\langle #1,#2 \rangle}

\usepackage[pagebackref=true,breaklinks=true,letterpaper=true,colorlinks,bookmarks=false]{hyperref}

\cvprfinalcopy

\begin{document}
\newcommand{\rg}{\mathbf{rk}}
\newcommand{\fA}{f_{\mbs \Theta_A}}
\newcommand{\fB}{f_{\mbs \Theta_B}}
\newcommand{\tA}{\mbs \Theta_A}
\newcommand{\tB}{\mbs \Theta_B}

\newcommand{\mbf}[1]{\ensuremath{\mathbf{#1}}}
\newcommand{\mbs}[1]{\ensuremath{\boldsymbol{#1}}}
\newcommand{\mcl}[1]{\ensuremath{\mathcal{#1}}}
\newcommand{\mrm}[1]{\ensuremath{\mathrm{#1}}}
\newcommand{\mbb}[1]{\ensuremath{\mathbb{#1}}}
\newcommand{\msf}[1]{\ensuremath{\mathsf{#1}}}

\newcommand{\ve}[1]{\ensuremath{\mathbf{#1}}} 
\newcommand{\m}[1]{\ensuremath{\mathsf{#1}}} 
\newcommand{\s}[1]{\ensuremath{\mathcal{#1}}} 

\newcommand{\trsp}[1]{\ensuremath{#1^{\top}}}
\newcommand{\pinv}[1]{\ensuremath{#1^{\dagger}}}
\newcommand{\bmat}[4]{\ensuremath{\begin{bmatrix}#1&#2\\#3&#4\end{bmatrix}}}
\def\trace{\ensuremath{\mathrm{trace}}}
\def\deter{\ensuremath{\mathrm{det}}}
\def\diag{\ensuremath{\mathrm{diag}}}
\def\rank{\ensuremath{\mathrm{rank}}}
\def\Id{\m{Id}} 
\def\mA{\m{A}}
\def\mB{\m{B}}
\def\mC{\m{C}}
\def\mD{\m{D}}
\def\mE{\m{E}}
\def\mF{\m{F}}
\def\mG{\m{G}}
\def\mH{\m{H}}
\def\mK{\m{K}}
\def\mL{\m{L}}
\def\mN{\m{M}}
\def\mP{\m{P}}
\def\mW{\m{W}}
\def\mX{\m{X}}
\def\mY{\m{Y}}
\def\mZ{\m{Z}}

\newcommand{\bvec}[2]{\ensuremath{\begin{bmatrix}#1\\#2\end{bmatrix}}}
\def\One{\mbs{1}} 
\def\va{\ve{a}}
\def\vb{\ve{b}}
\def\vc{\ve{c}}
\def\vd{\ve{d}}
\def\vf{\ve{f}}
\def\vg{\ve{g}}
\def\vh{\ve{h}}
\def\vi{\ve{i}}
\def\vt{\ve{t}}
\def\bx{\ve{x}}
\def\by{\ve{y}}
\def\bz{\ve{z}}
\def\bv{\ve{v}}

\def\cL{\mcl{L}}

\def\ie{\emph{i.e.}}
\def\eg{\emph{e.g.}}
\def\iid{\emph{i.i.d.}}
\def\wrt{w.r.t.}
\def\mwrt{\mrm{w.r.t.}}
\def\msbt{\mrm{sb.t.}}

\def\sqt{^{\frac{1}{2}}} 
\def\msqt{^{-\frac{1}{2}}} 
\def\R{\mbb R}
\def\vtheta{\mbs{\theta}}

\title{SoDeep: a Sorting Deep net to learn ranking loss surrogates}

\author{Martin Engilberge\textsuperscript{1,2}, Louis Chevallier\textsuperscript{2}, Patrick P{\'e}rez\textsuperscript{3}, Matthieu Cord\textsuperscript{1,3}\\[1ex]
\textsuperscript{1}Sorbonne Universit\'e, Paris, France, 
\textsuperscript{2}Technicolor, Cesson S\'evign\'e, France,
\textsuperscript{3}Valeo.ai, Paris, France\\[1ex]
{\tt\small \{martin.engilberge, matthieu.cord\}@lip6.fr}  {\tt\small patrick.perez@valeo.com  louis.chevallier@technicolor.com}
}

\maketitle

\begin{abstract}
Several tasks in machine learning are evaluated using non-differentiable metrics such as mean average precision or Spearman correlation. However, their non-differentiability prevents from using them as objective functions in a learning framework. Surrogate and relaxation methods exist but tend to be specific to a given metric. 

In the present work, we introduce a new method to learn approximations of such non-differentiable objective functions. Our approach is based on a deep architecture that approximates the sorting of arbitrary sets of scores. It is trained virtually for free using synthetic data. This sorting deep (SoDeep) net can then be combined in a plug-and-play manner with existing deep architectures. We demonstrate the interest of our approach in three different tasks that require ranking: Cross-modal text-image retrieval, multi-label image classification and visual memorability ranking. Our approach yields very competitive results on these three tasks, which validates the merit and the flexibility of SoDeep as a proxy for sorting operation in ranking-based losses. 
\end{abstract}

\section{Introduction}\label{sec:intro}
﻿﻿
Deep learning approaches have gained enormous research interest for many Computer Vision tasks in the recent years. Deep convolutional networks are now commonly used to learn state-of-the-art models for visual recognition, including image classification \cite{krizhevsky2012imagenet,He2016,simonyan2014very} and visual semantic embedding \cite{Kiros2014,Karpathy2015,wang2018learning}. One of the strengths of these deep approaches is the ability to train them in an end-to-end manner removing the need for handcrafted features \cite{lowe1999object}. In such a paradigm, the network starts with the raw inputs, and handles feature extraction (low level and high-level features) and prediction internally. 
The main requirement is to define a trainable scheme. For deep architectures, stochastic gradient descent with back-propagation is usually performed to minimize an objective function. This loss function depends on the target task but has to be at least differentiable. 

\begin{figure}[t!]
 \begin{center}
  \includegraphics[width=1.0\linewidth]{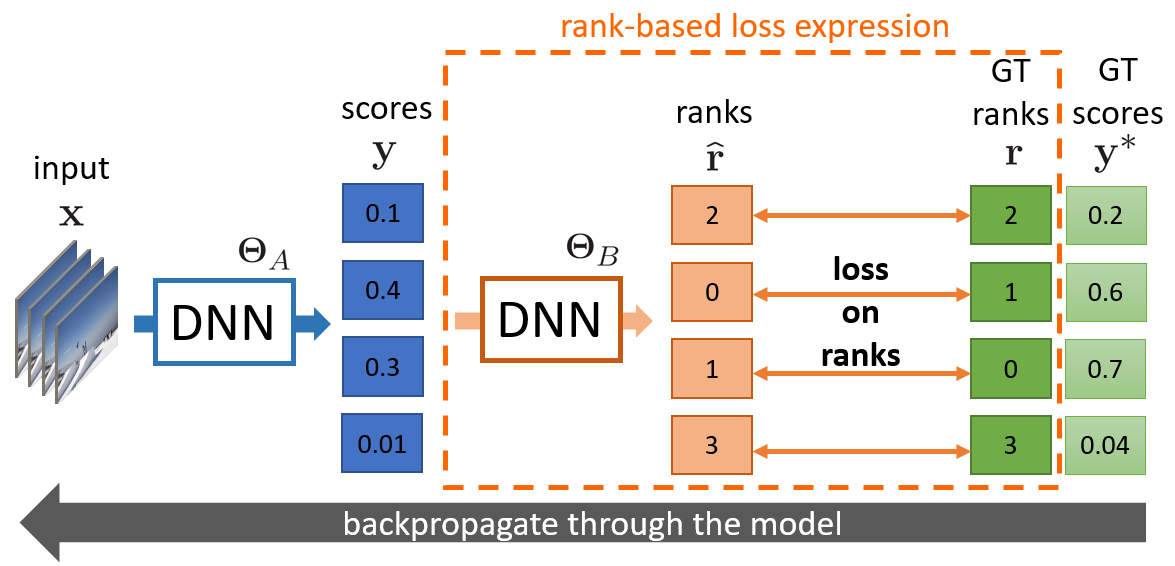}
 \end{center}
 \caption{\textbf{Overview of SoDeep, the proposed end-to-end trainable deep architecture to approximate non-differentiable ranking metrics}. A pre-trained differentiable sorter (deep neural net [DNN] $\Theta_B$) is used to convert into ranks the raw scores given by the model (DNN $\Theta_A$) being trained to a collection of inputs. A loss is then applied to the predicted rank and the error can be back-propagated through the differentiable sorter and used to update the weights $\Theta_A$. }
 \label{fig:intro_fig}
\end{figure}

Machine learning tasks are often evaluated and compared using metrics which differ from the objective function used during training. The choice of an evaluation metric is intimately related to the definition of the task at hand, even sometimes to the benchmark itself. For example, accuracy seems to be the natural choice to evaluate classification methods, whereas the choice of the objective function is also influenced by the mathematical properties that allow a proper optimization of the model. For classification, one would typically choose the cross entropy loss -- a differentiable function -- over the non-differentiable accuracy.
Ideally, the objective function used during training would be identical to the evaluation metric. However, standard evaluation metrics are often not suitable as training objectives for lack of differentiability to start with. This results in the use of surrogate loss functions that are better behaved (smooth, possibly convex). Unfortunately, coming up with good surrogate functions is not an easy task. 
\vspace{1mm}

In this paper, we focus on the non-differentiability of the evaluation metrics used in ranking-based tasks such as recall, mean average precision and Spearman correlation. Departing from prior art on building surrogates losses for such tasks, we adopt a simple, yet effective, learning approach: Our main idea is to approximate the non-differentiable part of such ranking-based metrics by an all-purpose learnable deep neural network. In effect, this architecture is designed and trained to mimic sorting operations. We call it SoDeep.  
SoDeep can be added in a plug-and-play manner on top of any deep network trained for tasks whose final evaluation metric is rank-based, hence not differentiable. The resulting combined architecture is end-to-end learnable with a loss that relates closely to the final metric. 

Our contributions are as follows:
\vspace{-1mm}
\begin{itemize}[leftmargin=1em]
\setlength\itemsep{0em}
    \item We propose a deep neural net that acts as a differentiable proxy for ranking, allowing one to rewrite different evaluation metrics as functions of this sorter, hence making them differentiable and suitable as training loss.
    \item We explore two types of architectures for this trainable sorting function: convolutional and recurrent. 
    \item We combine the proposed differentiable sorting module with standard deep CNNs, train them  end-to-end on three challenging tasks, and demonstrate the merit of this novel approach through extensive evaluations of the resulting models. 
\end{itemize}
The rest of the paper is organized as follows. We discuss in Section \ref{sec:related} the related works
on direct and indirect optimization of ranking-based metrics, and position our work accordingly. 
Section \ref{sec:approach} is dedicated to the presentation of our approach. We show in particular how a ``universal'' sorting proxy suffices to tackle standard rank-based metrics, and present different architectures to this end. 
More details on the system and its training are reported in Section \ref{sec:expe}, along with various experiments. We first establish new state-of-the-art performance on cross-modal retrieval, then we show the benefits of our learned loss function compared to standard methods on memorability prediction and multi-label image classification. 

\section{Related works}\label{sec:related}
Many data processing systems rely on sorting operations at some stage of their pipeline. It is the case also in machine learning, where handling such non-differentiable, non-local operations can be a real challenge \cite{mohri}. For example, retrieval systems require to rank a set of database items according to their relevance to a query. For sake of training, simple loss functions that are decomposable over each training sample have been proposed as for instance in \cite{Herschtal2004} for  the area under the ROC curve. Recently, some more complex non-decomposable losses (such as the Average Precision (AP), Spearman coefficient,   and normalized discounted cumulative gain (nDCG)
\cite{chakrabarti2008structured}) that present hard computational challenges have been proposed \cite{mohapatra2018efficient}.

\paragraph{Mean average precision optimization} 
Our work shares the high level goal of using ranking metrics as training objective function with many works before us. Several works studied the problem of optimizing average precision with support vector machines  \cite{joachims2002optimizing,yue2007support} and other works extended these approaches to neural networks \cite{burges2005learning, mohapatra2018efficient, dubey2016deep}. 
To learn rank, the seminal work \cite{joachims2002optimizing}  relies on a structured hinge upper bound to the loss. Further works reduce the computational complexity \cite{mohapatra2018efficient} or rely on asymptotic methods \cite{song2016training}.
The focus of these works is mainly on the relaxation of the mean average precision, while our focus is on learning a surrogate for the ranking operation itself such that it can be combined with multiple ranking metrics.
In contrast to most ranking-based techniques, which have to face the high computational complexity of the loss augmented inference \cite{joachims2002optimizing, song2016training, mohapatra2018efficient}, we propose a fast, generic, deep sorting architecture that can be used in gradient-based training for rank-based tasks.

\paragraph{Application of ranking based metrics} 
Ranking is commonly used in evaluation metrics. On retrieval tasks such as cross-modal retrieval \cite{Kiros2014, Karpathy2015, Frome2013, Faghri2017,Ma2015}, recall is the standard evaluation. Image classification \cite{everingham2007pascal, durand2017wildcat} and object recognition are evaluated with mean average precision in the multi-label case. Ordinal regression \cite{cohendet2018mediaeval} is evaluated using Spearman correlation.

\paragraph{Existing surrogate functions}  
Multiple surrogates for ranking exist. Using metric learning to do retrieval is one of them. This popular approach avoids the use of the ranking function altogether. Instead, pairwise \cite{xing2003distance}, triplet-wise \cite{weinberger09distance,chechik2010large} and list-wise \cite{fernando2015learning, cao2007learning} losses are used to optimize distances in a latent space. The cross-entropy loss is typically used for multi-label and multi-class classification tasks.

\section{SoDeep approach}\label{sec:approach}
Rank-based metrics such as recall, Spearman correlation and mean average precision can be expressed as a function of the rank of the output scores. The computation of the rank being the only non-differentiable part of these metrics, we propose to learn a surrogate network that approximates directly this sorting operation. 

\subsection{Learning a sorting proxy}

\begin{figure}[htb]
 \begin{center}
  \includegraphics[width=0.9\linewidth]{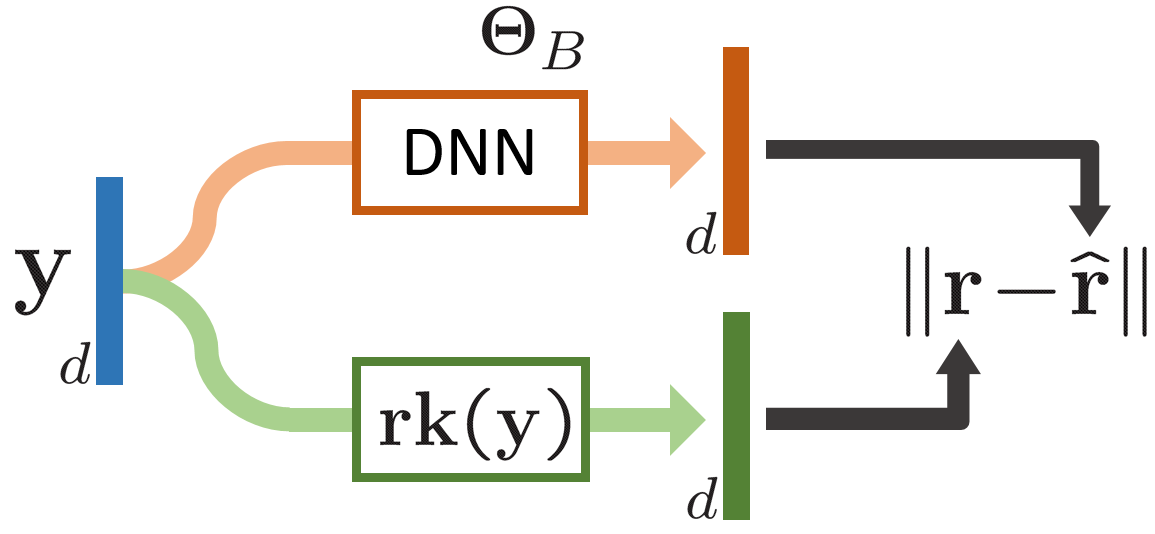}
 \end{center}
 \caption{\textbf{Training a differentiable sorter}. Given a score vector \ve{y} we learn the parameters $\tB$ of a DNN such that its output $\mathbf{\hat{r}}$ approximates
 the true rank vector $\rg(\ve y)$. The model is trained using gradient descent and an $L_1$ loss. Once trained, $\fB$ can be used as a differentiable surrogate of the ranking function.}
 \label{fig:diff_ranker}
\end{figure}

Let $\ve y\in\mathbb{R}^d$ be a vector of $d$ real values and $\rg$ the ranking function so that $\rg(\ve y)\in \{1 \cdots d\}^d$ is the vector containing the rank for each variable in $\ve y$, \ie~ $\rg(\ve y)_i$ is the rank of $\ve y_i$ among the $\ve y_j$'s. We want to design a deep architecture $\fB$ that is able to mimic this sorting operator.
The training procedure of this DNN is summarized in Fig. \ref{fig:diff_ranker}. The aim is to learn its parameters, $\tB$, so that the output of the network is as close as possible to the output of the exact sorting. 

Before discussing possible architectures, let's consider the training of this network, independent of its future use. We first generate a training set by randomly sampling $N$ input vectors $\ve y^{(n)}$ and we compute through exact sorting the associated ground-truth rank vectors $\ve r^{(n)}=\rg(\ve y^{(n)})$. We then classically learn the DNN $\fB$ by minimizing a $L_1$ loss between the predicted ranking vector $\mathbf{\hat{r}}=\fB(\ve y)$ and the ground-truth rank $\ve r$ over the training set:
\begin{equation}
    \min_{\tB} \sum_{n=1}^N \left\|\rg(\ve y^{(n)}) -  \fB(\ve y^{(n)})\right\|_1.
    \label{eq:sodeep_train}
\end{equation}
We explore in the following different network architectures and we explain how the training data is generated. 

\subsubsection{Sorter architectures}

We investigate two types of architectures for our differentiable sorter $\fB$. One is a recurrent network and the other one a convolutional network, each capturing interesting aspects of standard sorting algorithms:
\begin{figure*}[t!]
    \begin{subfigure}[t]{0.5\textwidth}
        \centering
        \includegraphics[width=0.85\textwidth]{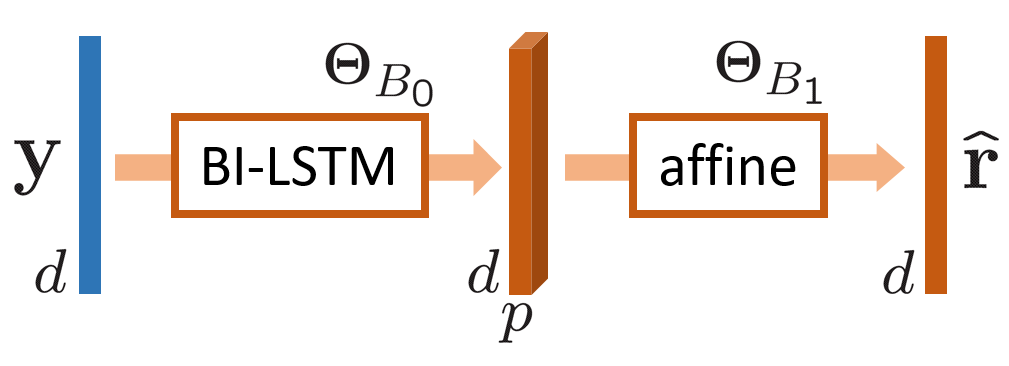}
        \caption{\textbf{Architecture of LSTM sorter}. }
        \label{fig:rank_archi_lstm}
    \end{subfigure}\hfill
    \begin{subfigure}[t]{0.5\textwidth}
        \centering
        \includegraphics[width=0.99\textwidth]{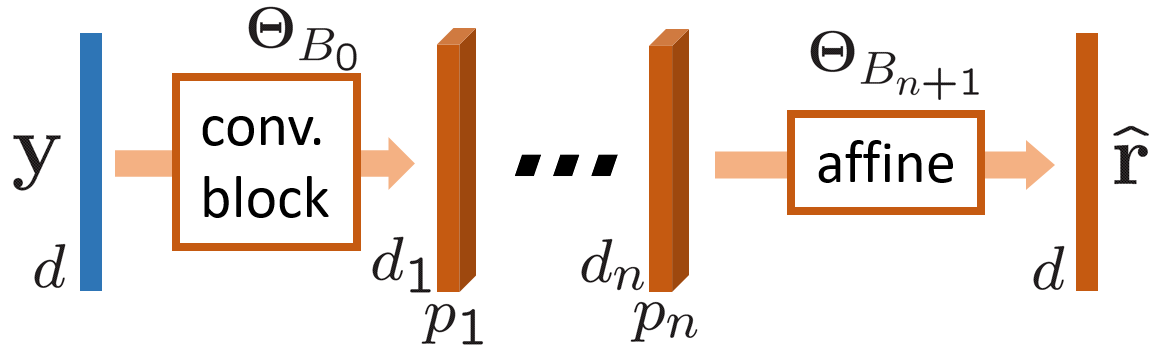}
        \caption{\textbf{Architecture of CNN sorter}.}
        \label{fig:rank_archi_cnn}
    \end{subfigure}
    \caption{\textbf{SoDeep architecture.} The sorter takes a vector of raw score $\ve y\in\R^{d}$ as input and outputs a vector $\mathbf{\hat{r}} \in \R^{d}$. Two architectures are explored, one recurrent (a), the other one, convolutional (b). Both architectures present a last affine layer to get a final projection to a vector $\mathbf{\hat{r}} $ in $ \R^{d}$. Note that even if it is not explicitly enforced, $\mathbf{\hat{r}}$ will try to mimic as close as possible the vector of the ranks of the $\ve y$ variables.}
    \label{fig:rank_archi} 
\end{figure*}
\begin{itemize}[leftmargin=1em]
\setlength\itemsep{0em}
    \item 
The recurrent architecture in Fig. \ref{fig:rank_archi_lstm} consists of a bi-directional LSTM \cite{schuster1997bidirectional} followed by a linear projection. The bi-directional recurrent network creates a connection between the output of the network and every input, which is critical for ranking computation. Knowledge about the whole sequence is needed to compute the true rank of any element.
    \item 
    The convolutional architecture in Fig. \ref{fig:rank_archi_cnn} consists of 8 convolutional blocks, each of these blocks being a one-dimensional convolution followed by a batch normalization layer \cite{ioffe2015batch} and a ReLU activation function. The sizes of the convolutional filters are chosen such that the output of the network contains as many channels as the length of the input sequence. Convolutions are used for their local property: indeed, sorting algorithms such as bubble sort \cite{friend1956sorting} only rely on a sequence of local operations. The intuition is that a deep enough convolutional network, with its cascaded local operations, should be able to mimic recursive sorting algorithms and thus to provide an efficient approximation of ranks.
\end{itemize}
We will further discuss  the interest of both types of SoDeep block architectures in the experiments. 

\subsubsection{Training data}\label{sec:training_data}
SoDeep module can be easily (pre)trained with supervision on synthetic data. Indeed, while being non-differentiable, the ranking function $\rg$ can be computed with classic sorting algorithms. 
The training data consists of vectors 
of randomly generated scalars, associated with their ground-truth rank vectors.  
In our experiments, the numbers are sampled from different types of distributions: 
\begin{itemize}[leftmargin=1.em]
\setlength\itemsep{0em}
\item Uniform distribution over $[-1,1]$; 
\item Normal distribution with $\mu = 0$ and $\sigma = 1$;
\item Sequence of evenly spaced numbers in a uniformly drawn random sub-range of $[-1,1]$;
\item Random mixtures of the previous distributions. 
\end{itemize}

While the differentiable sorter can be trained ahead of time on a variety of input distributions, as explained above, there might be a shift with the actual score distribution that the main network $\fA$ will output for the task at hand. 
This shift can reduce naturally during training, or an alignment can be explicitly enforced. For example, $\fA$ can be designed to output data in the interval used to learn the sorter, with the help of bounded functions such as cosine similarity.

\subsection{Using SoDeep for training with rank-based loss}
Rank-based metrics are used for evaluating and comparing learned models in a number of tasks. Recall is a standard metric for image and information retrieval, mean Average Prediction (mAP) for classification and recognition, and Spearman correlation for ordinal prediction. This type of rank-based metrics are non-differentiable because they require to transition from the continuous domain (score) toward the discrete domain (rank). 

As presented in Fig \ref{fig:intro_fig}, we propose to insert a pre-trained SoDeep proxy block $\fB$  between the deep scoring function $\fA$  and the chosen rank-based loss. We show in the following how mAP, Spearman correlation and recall can be expressed as functions of the rank and combined with SoDeep accordingly.

In the following we assume a training set of annotated pairs $\{(\ve x_i,y_i^*)\}_{i=1}^M$ for the task at hand. A group $\mathcal{B}$ of $d$ training examples among them yields a prediction vector $\ve y(\tA)= [\fA(\ve x_i)]_{i \in \mathcal{B}}$ and an associated ground-truth score vector $\ve y^*= [y_i^*]_{i \in \mathcal{B}}$  (Fig. \ref{fig:intro_fig}).   

\subsubsection{Spearman correlation}

For two vectors $\ve y$ and $\ve y'$ of size $d$, corresponding to two sets of $d$ observations, the Spearman correlation \cite{dodge2008concise} is defined as:
\begin{equation}\label{eq:spearcorr}
   r_s = 1 - \frac{6\lVert \rg(\ve y) - \rg(\ve y')\rVert_2^2}{d(d^2-1)}.
\end{equation}

\textit{Maximizing} w.r.t. parameters $\tA$ the sum of Spearman correlations \eqref{eq:spearcorr} between ground truth and predicted score vectors over $N$ subsets of training examples amounts to solving the minimization problem:
\begin{equation}
    \min_{\tA}  \sum_{n=1}^N\left\|\rg(\ve y^{(n)}) - \rg(\ve y^{*(n)}) \right\|^2_2,
\end{equation}
with the loss not being differentiable.

Using now our differentiable proxy instead of the rank function, we can define the new Spearman loss for a group $\mathcal B$:
\begin{equation}\label{eq:loss_spearcorr}
    \mathcal{L}_{SPR}(\tA, \mathcal{B}) = 
    \sum_{n=1}^N\left\|\fB(\ve y(\tA)^{(n)}) - \rg(\ve y^{*(n)}) \right\|^2_2.
\end{equation}
Training will typically minimize it over a large set of groups. 
Note that here the optimization is done over $\tA$, knowing that SoDepp block $\fB$ has been trained independently on specific synthetic training data. Optionally, the block can be fine-tuned along the way, hence minimizing w.r.t. $\tB$ as well.

\subsubsection{Mean Average Precision (mAP)}

Multilabel image classification is often evaluated using mAP, a metric from information retrieval. To define it, each of the $C$ classes is considered as a query over the $d$ elements of the datasets. For class $c$, denoting $\ve y_c^*$ the $d$-dimensional ground-truth binary vector and $\ve y_c$ the vector of scores for this class, the average precision (AP) for the class is defined as \cite{yue2007support} :
\begin{equation}\label{eq:ap}
   AP(\ve y_c,\ve y^*_c) = \frac{1}{\mathrm{rel}}\sum_{j:\ve y^*_c(j)=1}Prec(j),
\end{equation}
where $\mathrm{rel} = |j:\ve y^*_c(j)=1|$ is the number of positive items for class $c$ and precision for element $j$ is defined as:
\begin{equation}\label{eq:map_prec}
   Prec(j) =  \frac{|\{s \in \mathcal{S}\,:\,\ve y_c^*(s)=1\}|}{\rg(\ve y_c)_j},
\end{equation}
with $\mathcal{S}$ the set of indices of the elements of $\ve y_c$ larger than $\ve y_{c}(j) $.

Minimizing $\rg(\ve y)_j$ for all $j$ from class $c$ (i.e., those verifying $\ve y^*_c(j) = 1$) will be used as a surrogate of the maximization of the AP over predictor's parameters $\tA$.

The mAP is obtained by averaging AP over the $C$ classes. Replacing the rank function by its differentiable proxy, the proposed mAP-based loss reads: 
\begin{equation}\label{eq:loss_map}
    \mathcal{L}_{mAP}(\tA, \mathcal{B}) = \sum_{c=1}^C \si{f_{\tB}(\ve y_c)}{\ve y^*_c}.
\end{equation}

\subsubsection{Recall at $K$}
Recall at rank $k$ is often used to evaluate retrieval tasks. 
In the following we assume a training set $\{\ve x_i\}_{i=1}^M$ for the task at hand. A group $\mathcal{B}$ of $d$ training examples among them yields a $d\times d$ prediction matrix $\ve Y(\tA)= [\fA(\ve x_i)]_{i \in \mathcal{B}}$ representing the scores of all pairwise combinations of training examples in $\mathcal{B}$. In other words, the $i$-th column of this matrix, $\ve Y[i] = \fA(\ve x_i)$, provides the relevance of other vectors in the group w.r.t. to query $\ve x_i$.

This matrix being given, recall at $K$ is defined as:
\begin{equation}\label{eq:recall}
  R@K(\ve Y) = \frac{1}{d} \sum_{i=1}^d \begin{cases}
    1,& \text{if } \rg(\ve Y[i])_p < K\\
    0,              & \text{otherwise,}
 \end{cases} 
\end{equation}
with $p$ the index of the unique positive entry in $\ve Y[i]$, a single relevant item being assumed for query $\ve x_i$. 

Once again, our sorter enables a differentiable implementation of this measure. However, we could not obtain conclusive results yet, possibly due to the batch size limiting the range of the summation. We found, however, an alternative way to leverage our sorting network.
It is based on the use of the ``triplet loss'', a popular surrogate for recall. 
We propose to apply this loss on ranks instead of similarity scores, making it only dependent on the ordering of the retrieved elements. 
The triplet loss on the rank can be expressed as follows:
\begin{equation}
\mathrm{loss}(\ve Y[i], p, c) = \max\big\{0,\alpha +  \fB(\ve Y[i])_p -  \fB(\ve Y[i])_c\big\}, 
\label{eq:triplet}
\end{equation}
where $p$ is defined as above (the positive example in the triplet, given anchor query $\ve x_i$) and $c$ is the index of a negative (irrelevant) example for this query.   
The goal is to minimize the rank of the positive pair with score $\ve Y[i]_p$ such that its rank is lower than the rank of the negative pair with score $\ve Y[i]_c$ by a margin of $\alpha$.

The complete loss is then expressed over all the elements of $\mathcal{B}$ in its \textit{hard} negative version  as:
\begin{equation}\label{eq:loss_recall}
\mathcal{L}_{REC}(\tA, \mathcal{B}) =  
\frac{1}{d}\sum_{i\in \mathcal{B}} \max_{c\neq p,c\neq i}  \mathrm{loss}(\ve Y[i], p, c).
\end{equation}  

\paragraph{}
In equations \eqref{eq:spearcorr}, \eqref{eq:ap} and \eqref{eq:recall}, the metrics are expressed in function of the non-differentiable rank function $\rg$. Leveraging our differentiable surrogate allows us to design a differentiable loss function for each of these metrics, respectively \eqref{eq:loss_spearcorr}, \eqref{eq:loss_map} and \eqref{eq:loss_recall}.

\section{Experiments}\label{sec:expe}
We present in this section several experiments to evaluate our approach. We first detail the way we train our differentiable sorter deep block using only synthetic data.  We also  present a comparison between the different models based on CNNs and on LSTM recurrent nets and with our baseline inspired from pairwise comparisons. 
We then evaluate the SoDeep combined with deep scoring functions $f_{\mbs \Theta_B}$. The loss functions expressed in \eqref{eq:loss_spearcorr}, \eqref{eq:loss_map} and \eqref{eq:loss_recall} are applied to three different tasks: memorability prediction, cross-modal retrieval, and object recognition. 

\subsection{SoDeep Training and Analysis}

\subsubsection{Training}
The proposed SoDeep models based on BI-LSTM and CNNs are trained on synthetic pairs of scores and ranks generated on the fly according to the distributions defined in Section \ref{sec:training_data}.

For convenience we  call an epoch as going through 100 000 pairs. The training is done using the Adam optimizer \cite{kinga2015method} with a learning rate of $0.001$ which is halved every 100 epochs. Mini-batches of size 512 are used. The model is trained until the loss values stop decreasing and are stable.

\subsubsection{A handcrafted sorting baseline}
We add to our trainable SoDeep blocks a baseline that does not require any training.

Inspired by the representation of the ranking problem as a matrix of pairwise ordering in \cite{yue2007support}, we build a handcrafted differentiable sorter $f_{h}$ using pairwise comparisons.

A \textit{sigmoid} function parametrized with $\lambda$ scalar is used as a binary comparison function between two scalars $a$ and $b$ as:
\begin{equation}\label{eq:comp_hand}
\sigma_{comp}(a,b) =  \frac{\mathrm{1} }{\mathrm{1} + e^{-\lambda (b-a) }}.  
\end{equation}
Indeed, if $a$ and $b$ are separated by a sufficient margin, $\sigma_{comp}(a,b)$ will be either $0$ or $1$. The parameter $\lambda$ is used to control the precision of the comparator.

This function may be used to approximate the relative rank of two components $\ve{y}_i$ and $\ve{y}_j$ in a vector $\ve{y}$: $\sigma_{comp}(\ve{y}_i,\ve{y}_j)$ will be close to $1$ if $\ve{y}_i$ is (significantly) smaller than $\ve{y}_j$, 0 otherwise. 
By summing up the result of the comparison between $\ve{y}_i$ and all the other elements of the vector $\ve{y}$, we form
our ranking function $f_{h}$.
More precisely, the rank $f_{h}(\ve{y}, i)$ for the $i$-est element of $\ve y$ is expressed as follow:
\begin{equation}\label{eq:rank_hand}
   f_{h}(\ve{y}, i) = \sum_{j:j\neq i} \sigma_{comp}(\ve{y}_i,\ve{y}_j).
\end{equation}

The overall precision of the handcrafted sorter can be controlled by the hyper parameter $\lambda$. The value of lambda is a trade off between the precision of the predicted rank and the efficiency when back-propagating through the sorter. Further experiments will use $\lambda=10$. 

\subsubsection{Results}
Table \ref{tab:ranker_comp} contains the loss values of the two different trained sorters and the handcrafted one on a generated test set of 10 000 samples. The LSTM based sorter is the most efficient, outperforming the CNN and the handcrafted sorters.

\begin{table}[t]
\begin{center}
\rowcolors{2}{gray!20}{white}
\begin{tabular}{lcc}\rowcolor{gray!40}
Sorter model & L1 loss \\
Handcrafted sorter & 0.0350 \\
CNN sorter & 0.0120 \\
LSTM sorter loss  & \textbf{0.0033} \\

\end{tabular}
\end{center}
\caption{\textbf{Performance of the sorters on synthetic data}. Ranking performance of the sorter on the synthetic dataset. Among the learned sorters the LSTM one is the most efficient.}
\label{tab:ranker_comp}
\end{table}

The performance of the CNN sorter slightly below the LSTM-based one can be explained by local behaviour of the CNNs, requiring a more complex structure to be able to rank elements.

\begin{figure}[htb]
 \begin{center}
  \includegraphics[width=1.0\linewidth]{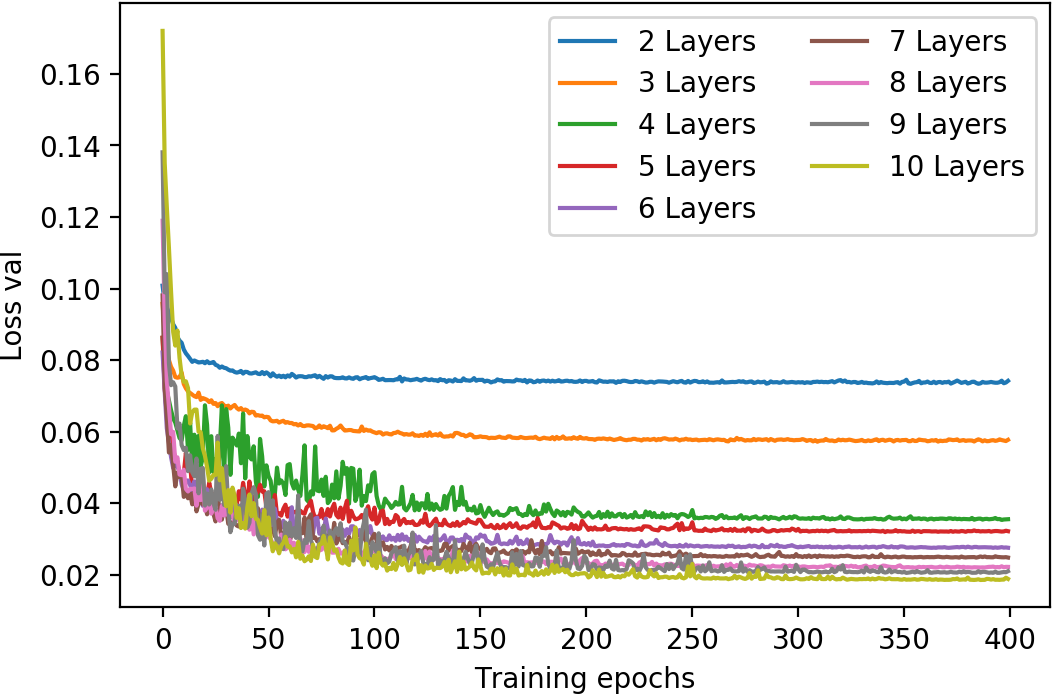}
 \end{center}
 \caption{\textbf{Performance of the CNN sorter with respect to the depth of the CNN}. Value of the cost function during the training of multiple CNN sorters with a number of layers varying from 2 to 10. The model performances saturate for models with 8 layers or more. }
 \label{fig:ranker_cnn_depth}
\end{figure}

\paragraph{}
In Figure \ref{fig:ranker_cnn_depth} we compare CNN sorters with respect to their number of layers. From these results, we choose to use 8 layers in our CNN sorter since the performance seems to saturate once this depth has been reached. A possible explanation of this saturation is that the relation between the depth of the network and the input dimension ($d=100$ here) is logarithmic.

\subsubsection{Further analysis}
\begin{figure}[htb]
 \begin{center}
  \includegraphics[width=0.9\linewidth]{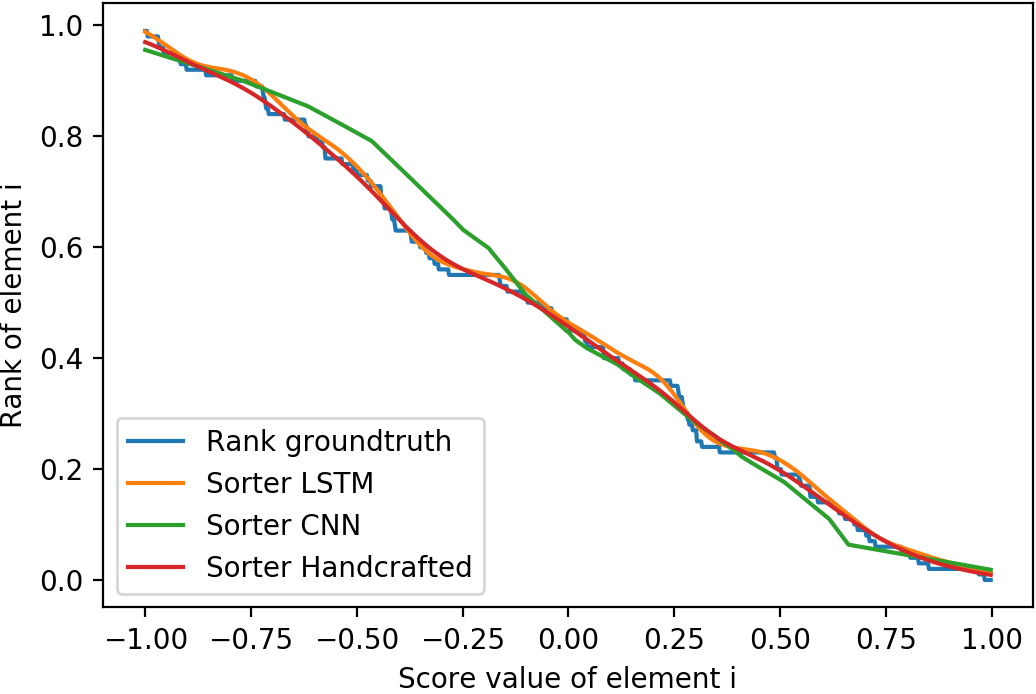}
 \end{center}
 \caption{\textbf{Sorter behaviour analysis}. Given a synthetic vector $\ve y'$ of raw scores in the range $[-1,1]$ of size 100 we plot the rank of its first element $\ve y'_1$ when the said value is linearly interpolated between -1 and 1. The x-axis represent the value $\ve y'_1$, and the y-axis is it corresponding rank. }
 \label{fig:ranker_swip}
\end{figure}

The ranking function being non-continuous is non-differentiable, the rank value is jumping from one discrete value to another. We design an experiment to visualize how the different types of sorter behave at these discontinuities. Starting from a uniformly sampled vector $\ve y' \in \R^{100}$ of raw scores in the range $[-1,1]$, we compute the ground truth rank $\rg(\ve y')_1$ and the predicted rank $\fB(\ve y')_1$ of the first element $y'_1$ while varying this element $y'_1$ from -1 to 1 in increments of 0.001.
The plot of the predicted ranks can be found in Fig. \ref{fig:ranker_swip}. The blue curve corresponds to the ground-truth rank where non-continuous steps are visible, whereas the curves for the learned sorters (orange and green) are a smooth approximation of the ground-truth curve. 

\begin{figure}[htb]
 \begin{center}
  \includegraphics[width=1.0\linewidth]{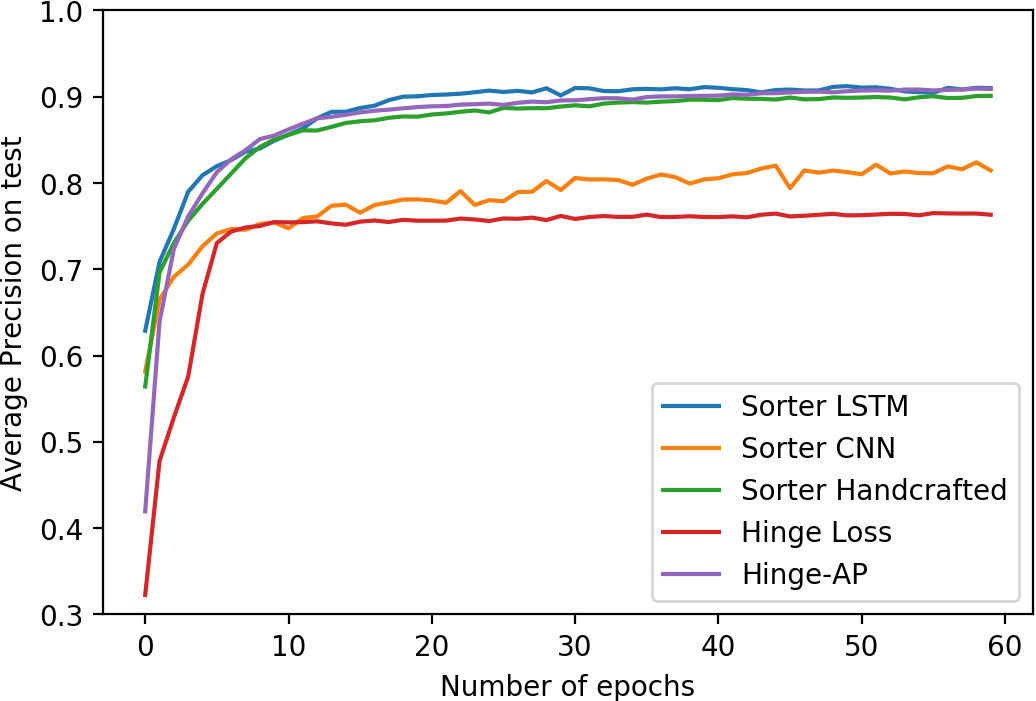}
 \end{center}
 \caption{\textbf{Synthetic experiment on mAP optimization}. Comparison against the proposed sorter and the previous approaches.}
 \label{fig:ranker_toy}
\end{figure}

In Fig. \ref{fig:ranker_toy} we compare our SoDeep against previous approaches optimizing structured hinge upper bound to the mAP loss. We followed the protocol described in \cite{song2016training} for their synthetic data experiments. 
Our sorters using the loss $\mathcal{L}_{mAP}$ defined in \eqref{eq:loss_map} are compared to a re-implementation of the Hinge-AP loss proposed in \cite{joachims2002optimizing}.
The results in Fig. \ref{fig:ranker_toy} show that our approach with the LSTM sorter (blue curve) gets mAP scores similar to \cite{joachims2002optimizing} (purple curve) while being generic and less complex.

\paragraph{}
From the learned sorters, the LSTM architecture is the one performing best on synthetic data (Tab. \ref{tab:ranker_comp}). In addition, its simple design and small number of hyper-parameters make it straightforward to train. The CNN architecture while not being as efficient, uses a smaller number of weights and is 1.7 time faster. Further experiments will use the LSTM sorter unless specified otherwise.

\subsection{Differentiable Sorter based loss functions}
Our method is benchmarked on three tasks. Each one of these tasks focuses on a different rank based loss function. Cross-modal retrieval will be used to test recall evaluation metrics, memorability prediction will be used for Spearman correlation and image classification will be used for mean average precision.

As explained in Section \ref{sec:training_data}, a shift in distribution might appear when using sorter-based loss. To prevent this, a parallel loss can be used to help domain alignment. This loss can be used only to stabilize the initialization or kept for the whole training.

\subsubsection{Spearman Correlation: Predicting Media Memorability}
\begin{figure}[htb]
 \begin{center}
  \includegraphics[width=1.0\linewidth]{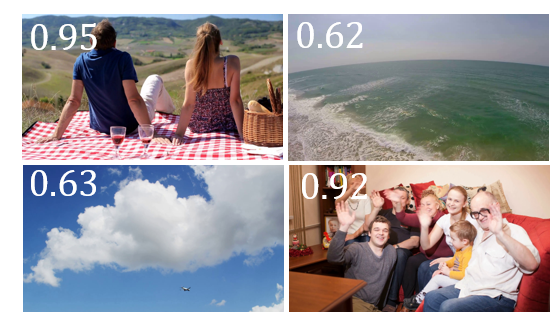}
 \end{center}
 \caption{\textbf{Media memorability dataset}. Frames with low and high memorability scores coming from 4 different videos of the memorability dataset \cite{cohendet2018mediaeval}. The memorability scores are overlayed on top of the images. }
 \label{fig:memo_dset}
\end{figure}

The media memorability prediction task \cite{cohendet2018mediaeval} is used to test the differentiable sorter with respect to the Spearman correlation metrics. Examples of elements of the dataset can be found in Fig. \ref{fig:memo_dset}.
Given a 7 seconds video the task consists in predicting the short term memorability score. The memorability score reflects the probability of a video being remembered. 

The task is originally on video memorability. However the model used here are pretrained on images, therefore 7 frames are extracted from each video and are associated with the memorability score of the source video. The training is done on pairs of frame and memorability score. During testing the predicted score of the 7 frames of a video are averaged to obtain the score per video. The dataset contains 8000 videos (56000 frames) for training and 2000 videos for testing. This training set is completed using LaMem dataset \cite{ICCV15_Khosla} adding 60 000 (image, memorability) pairs to the training data.

\begin{table}[t]
\begin{center}
\rowcolors{2}{gray!20}{white}
\begin{tabular}{lcc}\rowcolor{gray!40}
Single model & Spear. cor. test \\
Baseline \cite{cohendet2018transfer} & 46.0 \\
Image only \cite{gupta2018mediaeval} &  48.8 \\
R34 + MSE loss & 44.2 \\
R34 + SoDeep loss  & 46.6 \\
Sem-Emb + MSE loss & 48.6 \\
Sem-Emb + SoDeep loss & \textbf{49.4} \\
\end{tabular}
\end{center}
\caption{\textbf{Media Memorability prediction results}. Our proposed loss function and architecture outperform the state-of-the-art system \cite{gupta2018mediaeval} by 0.6 pt.}
\label{tab:memorability}
\end{table}

\paragraph{Architectures and training} The regression model consists of a feature extractor combined with a two layers MLP \cite{rosenblatt1958perceptron} regressing features to a single memorability score. We use two pretrained nets to extract features: the Resnet-34 \cite{He2016} and the semantic embedding model of \cite{engilberge2018finding} (as in the next section).

We use the loss $\mathcal{L}_{SPR}$ defined in \eqref{eq:loss_spearcorr} to learn the memorability model.  The training is done in two steps. First, for 15 epochs only the MLP layers are trained while the weights of the feature extractor are kept frozen. Second, the whole model is finetuned. The Adam optimizer \cite{kinga2015method} is used  with a learning rate of $0.001$ which is halved every 3 epochs. 
To help with domain adaptation, our loss is combined with an L1 loss for the first epoch.

\paragraph{Results} In Tab. \ref{tab:memorability}, we compare the impact of the learned loss over two architectures. For both models we defined a baseline using a L2 loss. On both architectures the proposed loss function achieves higher Spearman correlation by 2.4 points on the Resnet model and 0.8 points on the semantic embedding model. These are state of the arts result on the task with an absolute gain of 0.6 pt. The model is almost on par (-0.3 pt) with an ensemble method proposed by \cite{gupta2018mediaeval} that is using additional textual data.

\paragraph{Sorter comparison} The memorability prediction is also used to compare the different types of sorters presented so far. Fixing the model and the hyper parameters, 4 models are trained with 4 different types of loss. The losses based on the LSTM sorter, the CNN sorter and the handcrafted sorter obtained respectively a Spearman correlation of 49.4, 46.6, 45.7, and the L1 loss gives a correlation of 46.2. These results are consistent with the result on synthetic data, with the LSTM sorter performing the best, followed by the CNN and handcrafted ones.

\begin{table*}[ht]
\begin{center}
  \rowcolors{2}{gray!20}{white}
\begin{tabular}{r c c c c c  c c c c c } \rowcolor{gray!40}
 & & \multicolumn{4}{c}{caption retrieval} & & \multicolumn{4}{c}{image retrieval} \\ \rowcolor{gray!40}
model && R@1 & R@5 & R@10 & Med. r && R@1 & R@5 & R@10 & Med. r \\
Emb. network \cite{wang2018learning} && 54.9 & 84.0 & 92.2 & - && 43.3 & 76.4 & 87.5 & - \\
DSVE-Loc \cite{engilberge2018finding} && 69.8 & 91.9 & 96.6 & 1  && 55.9 & 86.9 & 94.0 & 1 \\
GXN (i2t+t2i) \cite{gu2018look} && 68.5 & - & \textbf{97.9} & 1  && \textbf{56.6} & - & \textbf{94.5} & 1 \\
DSVE-Loc + SoDeep loss && \textbf{71.5} & \textbf{92.8} & 97.1 & 1  && 56.2 & \textbf{87.0} & 94.3 & 1
\end{tabular}
\end{center}
  \caption{\textbf{Cross-modal retrieval results on \coco.} Using the proposed rank based loss function outperforms the hard negative triplet margin loss, achieving state-of-the-art results on the caption retrieval task. }  
 \label{tab:retrieval}
\end{table*}

\subsubsection{Mean Average precision: Image classification}

 The VOC 2007 \cite{everingham2007pascal} object recognition challenge is used to evaluate our sorter on a task using the mean average precision metric. We use an off-the-shelf model \cite{durand2017wildcat}. This model is a fully convolutional network, combining a Resnet-101 \cite{He2016} with advanced spatial aggregation mechanisms.
 
 To evaluate the loss $\mathcal{L}_{mAP}$ defined in \eqref{eq:loss_map} two versions of the model are trained: A baseline using only multi-label soft margin loss, and another model trained using the multi-label soft margin loss combined with $\mathcal{L}_{mAP}$.
 
 Rows 3 and 4 of  Tab.~\ref{tab:map_voc} show the results obtained by the two previously described models. Both models are below the state-of-the-art, however the use of the rank loss is beneficial and improves the mAP by 0.8 pt compared to the model using only the soft margin loss.

\begin{table}[t]
\begin{center}
\rowcolors{2}{gray!20}{white}
\begin{tabular}{lc}\rowcolor{gray!40}
Loss & mAP \\
VGG 16 \cite{simonyan2014very} & 89.3 \\
WILDCAT \cite{durand2017wildcat} & 95.0\\
WILDCAT*  & 93.2\\
WILDCAT* + SoDeep loss  & 94.0
\end{tabular}
\end{center}
\caption{\textbf{Object recognition results}. Model marked by (*) are obtained with code available online: \url{https://github.com/durandtibo/wildcat.pytorch} }
\label{tab:map_voc}
\end{table}

\subsubsection{Recall@K: Cross-modal Retrieval}

The last benchmark used to evaluate the differentiable sorter is the cross-modal retrieval. Starting from images annotated with text, we train a model producing rich features for both image and text that live in the same embedding space. Similarity in the embedding space is then used to evaluate the quality of the model on the cross-modal retrieval task.

Our approach is evaluated on the \coco~dataset~ \cite{lin2014microsoft} using the rVal split proposed in \cite{Karpathy2015}. The dataset contains 110k images for training, 5k for validation and 5k for testing. Each image is annotated with 5 captions.

Given a query image (resp. a caption), the aim is to retrieve the corresponding captions (resp. image). Since \coco~contains 5 captions per image, recall at $r$ (``R@$r$'') for caption retrieval is computed based on whether at least one of the correct captions is among the first $r$ retrieved ones. The task is performed 5 times on 1000-image subsets of the test set and the results are averaged.

We use an off-the-shelf model \cite{engilberge2018finding}. It is a two-paths multimodal embedding approach that leverages the latest neural network architecture. The visual pipeline is based on a Resnet-152 and is fully convolutional. The textual pipeline is trained from scratch and uses a Simple Recurrent Unit (SRU) \cite{Lei2017} to encode sentences. The model is trained using the loss $\mathcal{L}_{REC}$ defined in \eqref{eq:loss_recall} instead of the triplet based loss.

Cross-modal retrieval results can be found in Tab.~\ref{tab:retrieval}. The model trained using the proposed loss function (DSVE-Loc + SoDeep loss) outperforms the similar architecture DSVE-Loc trained with the triplet margin based loss by (1.7\%,0.9\%,0.5\%) on (R@1,R@5,R@10) in absolute for caption retrieval, and by (0.3\%,0.1\%,0.3\%) for image retrieval. It obtains state-of-the-art performance on caption retrieval and is very competitive on image retrieval being almost on par with the GXN \cite{gu2018look} model, which has a much more complex architecture. It is important to note that the loss function proposed could be beneficial for any type of architecture.

\section{Conclusion}\label{sec:conclusion}
We have presented SoDeep, a novel method that leverages the expressivity of recent architectures to learn differentiable surrogate functions. Based on a direct deep network modeling of the sorting operation, such a surrogate allows us to train, in an end-to-end manner, models on a diversity of tasks that are traditionally evaluated with rank-based metrics. Remarkably, this deep proxy to estimate the rank comes at virtually no cost since it is easily trained on purely synthetic data.

Our experiments show that the proposed approach achieves very good performance on cross-modal retrieval tasks as well as on media memorability prediction and multi-label image classification. These experiments demonstrate the potential and the versatility of SoDeep. This approach allows the design of training losses that are closer than before to metrics of interest, which opens up a wide range of other applications in the future.

{\small
\bibliographystyle{ieee}
\bibliography{egbib}
}

\end{document}